\documentclass[letterpaper, 10 pt, conference]{ieeeconf}

\IEEEoverridecommandlockouts                              

\usepackage{graphics} 
\usepackage{epsfig} 
\usepackage{mathptmx} 
\usepackage{times} 
\usepackage{amsmath} 
\usepackage{amssymb}  
\usepackage{cite}
\usepackage{algorithmic}
\usepackage[ruled,vlined]{algorithm2e}
\usepackage{color}
\usepackage{bm}
\usepackage{cite}
\usepackage{array}
\usepackage{soul}
\usepackage{color}
\usepackage{graphics} 
\usepackage{epsfig} 
\usepackage{mathptmx} 
\usepackage{times} 
\usepackage{amsmath} 
\usepackage{amssymb} 
\usepackage{hyperref}
\usepackage{float}
\usepackage{multirow}
\usepackage{paralist, tabularx}
\usepackage[dvipsnames]{xcolor}
\usepackage{soul}
\usepackage[normalem]{ulem}

\title{\LARGE \bf
Towards Terrain-Aware Safe Locomotion for Quadrupedal Robots Using Proprioceptive Sensing
}

\author{Peiyu Yang$^{+}$, Jiatao Ding$^{+,\star}$, Wei Pan, Claudio Semini, and Cosimo Della Santina
\thanks{Peiyu Yang and Claudio Semini are with the Dynamic Legged Systems, Istituto Italiano di Tecnologia (IIT), Italy (e-mail: peiyu.yang@iit.it; claudio.semini@iit.it). Jiatao Ding is with the Department of Industrial Engineering, University of Trento, Italy (e-mail: jiatao.ding@unitn.it). Wei Pan is with the Department of Computer Science, University of Manchester, UK (e-mail: wei.pan@manchester.ac.uk). Cosimo Della Santina is with Cognitive Robotics, Delft University of Technology, The Netherlands, and also with the Institute of Robotics and Mechatronics, German Aerospace Center (DLR), Germany (e-mail: c.dellasantina@tudelft.nl).
Peiyu Yang and Jiatao Ding contributed equally.
Jiatao Ding is the corresponding author. 
}}

\begin{document}

\maketitle
\thispagestyle{empty}
\pagestyle{empty}

\begin{abstract}

Achieving safe quadrupedal locomotion in real-world environments has attracted much attention in recent years. When walking over uneven terrain, achieving reliable estimation and realising safety-critical control based on the obtained information is still an open question. To address this challenge, especially for low-cost robots equipped solely with proprioceptive sensors (e.g., IMUs, joint encoders, and contact force sensors), this work first presents an estimation framework that generates a 2.5-D terrain map and extracts support plane parameters, which are then integrated into contact and state estimation. Then, we integrate this estimation framework into a safety-critical control pipeline by formulating control barrier functions that provide rigorous safety guarantees. Experiments demonstrate that the proposed terrain estimation method provides smooth terrain representations. Moreover, the coupled estimation framework of terrain, state, and contact reduces the mean absolute error of base position estimation by 64.8\%, decreases the estimation variance by 47.2\%, and improves the robustness of contact estimation compared to a decoupled framework. The terrain-informed CBFs integrate historical terrain information and current proprioceptive measurements to ensure global safety by keeping the robot out of hazardous areas and local safety by preventing body-terrain collision, relying solely on proprioceptive sensing.
\end{abstract}

\section{INTRODUCTION}
Quadrupedal robots have garnered significant attention in robotics research due to their high mobility and adaptability across unstructured environments\cite{panichi2025fly,atanassov2024curriculum,ding2024robust, hoeller2024anymal}. However, achieving safe locomotion---whether global safety by avoiding hazardous regions or local safety that prevents body-terrain collision---over uneven terrain remains a major challenge, particularly in demanding applications such as exploration and rescue in hazardous areas. To this end, terrain-aware estimation and safety-critical control have emerged as promising research directions\cite{gangapurwala2022rloc,shi2023terrain}.

In terms of terrain-aware estimation, the 2.5-dimensional (2.5-D) elevation map has been widely adopted for terrain representation due to its structured grid form\cite{wermelinger2016navigation,norby2022quad}. In existing work, the construction of 2.5-D maps with mobile robots typically relies on LiDARs or depth cameras. Some approaches, such as \cite{kweon1992high} and \cite{jiang2019simultaneous}, fuse multiple scans from exteroceptive sensors to generate a complete elevation map. \cite{belter2012estimating} and \cite{fankhauser2016universal} integrate the robots' motion poses with exteroceptive measurements to build local terrain maps that are dynamically updated during locomotion.
By combining proprioceptive and exteroceptive sensing modalities, the work in\cite{fankhauser2018probabilistic} proposed a novel probabilistic mapping method that integrates robot motion and range sensors, yielding a real-time, reliable estimation of terrain contours and confidence bounds. Additionally, based on terrain similarity, the work in \cite{xie2024real} proposed an adaptive fusion method to estimate the local map.
\begin{figure}
\centerline{\includegraphics[width=1\linewidth]{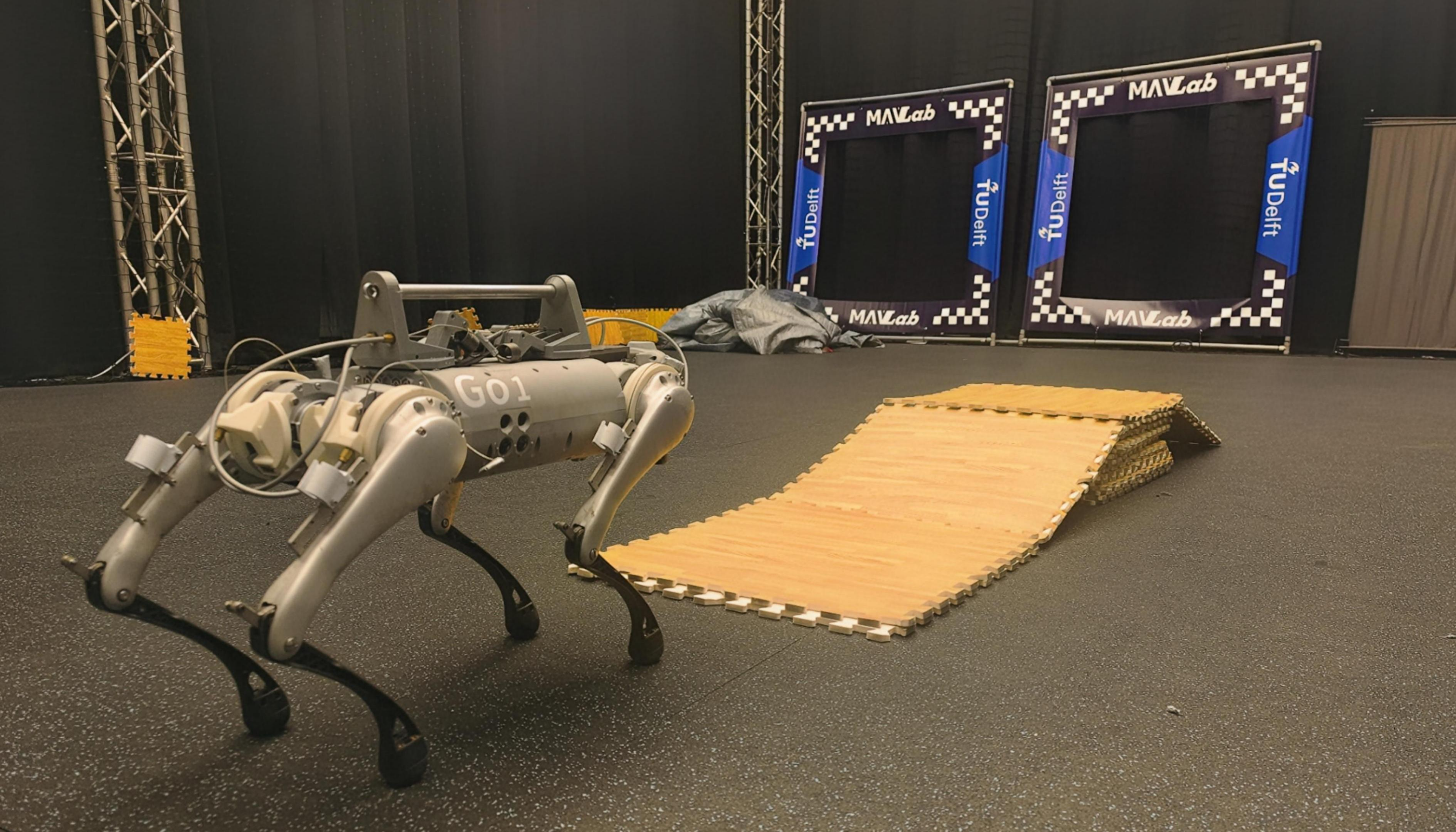}}
\vspace{-3mm}
\caption{An example of the scenario we deal with in this work. A quadruped is capable of walking on uneven terrain without any prior knowledge of the environment or any exteroceptive sensors, using the proposed method.}
\label{intro}\vspace{-0.6cm}
\end{figure}

However, the aforementioned methods rely on high-precision range sensors, which not only increase system cost and payload, but are also susceptible to occlusion, dust, and illumination variations, thereby limiting the stability and reliability under degraded perceptual conditions\cite{11045071, dhrafani2025firestereo}. As a result, proprioception-based terrain estimation has attracted increasing attention as a complementary solution. By leveraging contact sensing and state estimation, these methods directly infer terrain information even when exteroceptive perception is limited or unavailable. The local slope approximation method \cite{bledt2018cheetah,lin2020contact,wang2023estimation} uses four contact points to calculate plane parameters, but only captures local planes and lacks global terrain mapping. Yang et al.\cite{yang2023proprioception} proposed a 2.5-D mapping method using historical foothold positions to retain terrain features, but the required smoothing filter can blur terrain edges.
Several studies have examined terrain information in state\cite{yang2021a1qpmc} and contact\cite{bledt2018contact} estimation, but they either assumed a stochastic terrain model or used pre-known ground angles. Consequently, terrain estimation remains decoupled from the state and contact estimation process, limiting the integration of cross-domain information. To address these limitations, we propose a novel terrain estimation method based on probabilistic fusion that outputs both a smooth 2.5-D terrain map and plane parameters without post-processing. Furthermore, we realize coupled terrain, contact, and state estimation. 

Regarding motion control, control barrier functions (CBFs) have been widely adopted as an effective approach for achieving safe locomotion behaviors\cite{ding2025versatile, lee2024safety, shamsah2023integrated}. In contrast to learning-based safety control methods \cite{he2024agile,schneider2024learning}, CBFs offer rigorous safety guarantees \cite{brunke2022safe}.  By incorporating affine inequality constraints into a quadratic programming (QP) framework, the CBF-based approaches guarantee that the robot's state is confined within a forward-invariant safe set. Following this idea, \cite{grandia2021multi} combined CBFs with model predictive control (MPC)
for safe foot placement and stability, while \cite{unlu2024control} and \cite{saradagi2024body} addressed safe navigation in complex environments. 
Despite these advancements, existing studies remain dependent on exteroceptive sensors. In addition, safety is typically enforced at the task level, such as foot placement\cite{grandia2021multi}, climbing stairs\cite{li2023autonomous}, and navigation\cite{unlu2024control}, ignoring the local safety of body orientation that prevents collision and maintains stability.
From this perspective, we aim to take a step toward solving the above limitations by introducing a CBF-MPC framework that ensures both global and local safety using information estimated by the proposed terrain estimation module. 

To summarize, the main contributions of this work are:
\begin{itemize}
    \item  A terrain estimation approach based on probabilistic fusion that generates a 2.5-D map and plane parameters, together with a coupled terrain–contact–state estimation framework based solely on proprioceptive sensing.
    \item A CBF-MPC framework that leverages the estimated terrain information to provide guarantees for both global navigation safety and local body orientation safety.
    \item Experimental validation of the terrain estimation framework on the Unitree Go1 platform (see Fig.~\ref{intro}), together with extensive simulation studies validating the coupled estimation and the CBF-MPC framework, demonstrating safe locomotion over uneven terrain.
\end{itemize}

The remainder of this paper is organized as follows: Section \ref{sec:overview} provides an overview of the complete system. Section \ref{sec:est} details the estimation algorithm. Section \ref{sec:safe} introduces the CBF-MPC framework. Section \ref{results} presents the experimental results. Section \ref{Limitations} discusses the limitations and future work, and Section \ref{conclusion} concludes the paper.

\section{Overview of the Proposed Methodology}\label{sec:overview}

\begin{figure}[t]
\centerline{\includegraphics[width=\linewidth]{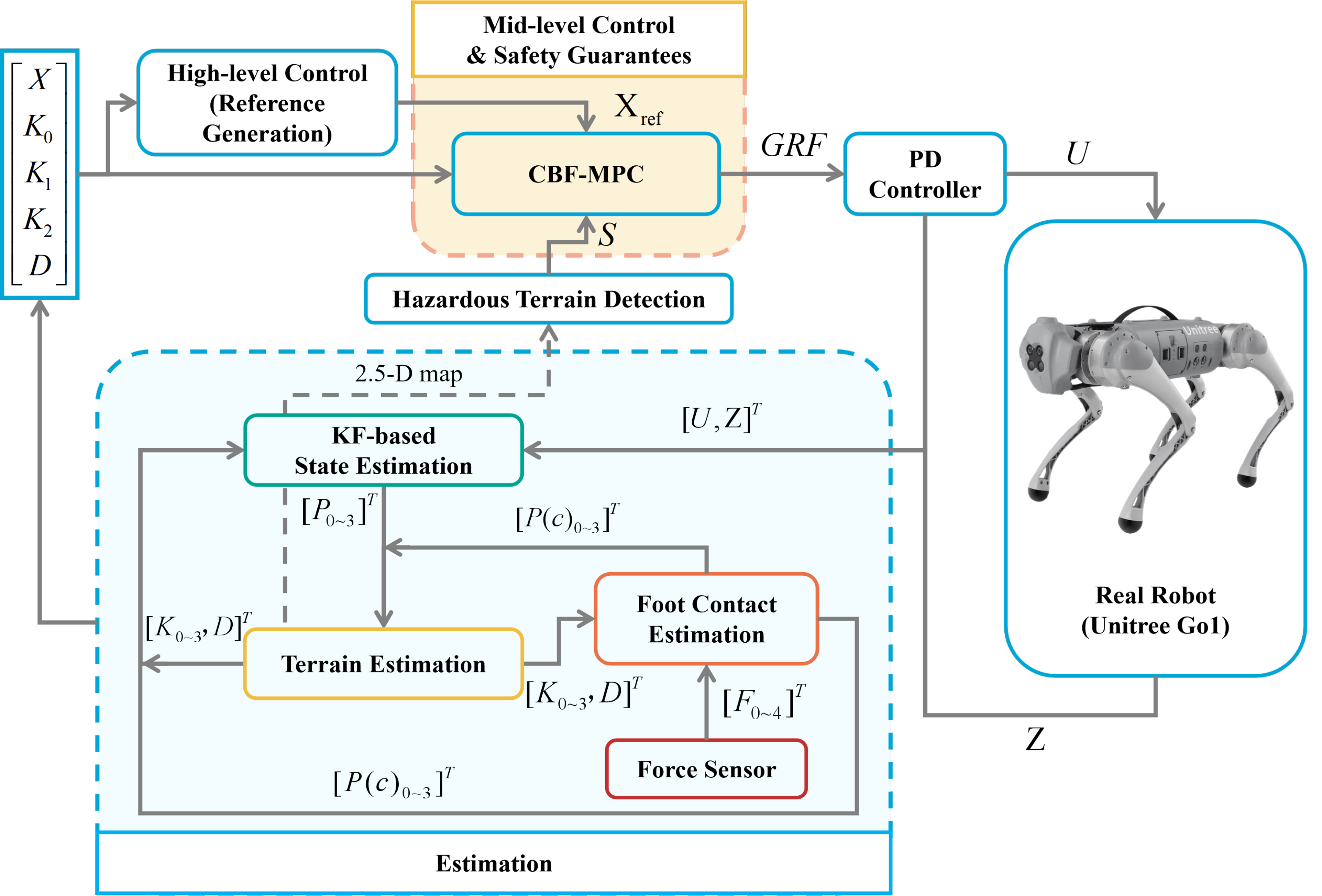}}
\vspace{-0.2cm}
\caption{The block diagram of the proposed control system. The blue shading zone shows the terrain-aware estimation system, and the yellow shading zone presents the CBF-MPC. In the figure, KF stands for the Kalman filter.}
\vspace{-3mm}
\label{overview}

\end{figure}

As depicted in Fig.~\ref{overview}, the proposed pipeline follows a hierarchical architecture, integrating high-level control, mid-level control, low-level PD control, and estimation, to achieve robust and safe locomotion for quadrupedal robots over uneven terrain. This work focuses on the estimation module (blue block) and the CBF-MPC module (yellow block).

In the estimation module, we establish a coupled estimation system that uses only IMU data, encoder data, and proportional values from force sensors. This system outputs the robot state $\bm{X}$, contact probabilities $P(c)_{i}$, a 2.5-D map, and plane parameters $K_1$, $K_2$, $K_3$, and $D$ that conform to the plane equation $K_1x+K_2y+K_3z+D=0$.

In the CBF-MPC module, we use the plane parameters and the hazardous point $\bm{S}$ to establish global and local CBFs constraints.

\textit{Notations:} In the following text, matrices and vectors are denoted using bold symbols (e.g., $\mathbf{A}$ for matrices and $\bm{v}$ for vectors). The notation $(\cdot)[k]$ refers to the $k$-th element of a vector or a matrix. Scalars are represented using regular italic symbols (e.g., $a$). Subscripts and superscripts are used to indicate specific indices or contexts, where necessary. In pseudocode, parameters are denoted using italic symbols (e.g., \textit{param}), while functions are represented using regular text (e.g., \textnormal{FunctionName}).

\section{Estimation}\label{sec:est}
This section details the proprioceptive-based estimation module, including terrain estimation, contact estimation, and state estimation.

\subsection{Terrain estimation}
The proposed terrain estimation method relies solely on proprioceptive measurements. Unlike the local slope approximation \cite{bledt2018cheetah} and foothold history filtering\cite{yang2023proprioception}, our algorithm obtains both a 2.5-D map and fitted plane parameters by probabilistically fusing the contact support areas during locomotion.

As shown in Algorithm \ref{TEalgorithm}, the procedure can be divided into three main phases: plane parameter computation, terrain map update, and local support plane computation. The inputs to this algorithm are: 1) \textit{Terrain}, representing the 2.5-D map; 2) \textit{P$_{contact}$}, which is the list of contact probability; and~3)~\textit{P$_{foot}$}, denoting the list of foot contact positions. 

In the first phase, for each leg, three points are selected: the position of the leg itself and those of its two neighboring legs, forming a triangle. The plane parameters of each triangle \textit{locPlane} is calculated via the $\operatorname{ComputePlane()}$ function:
\begin{equation}
    \begin{aligned}
        \begin{array}{c}  
            \begin{bmatrix}
                K_1 \\
                K_2 \\
                K_3
            \end{bmatrix}
            =
            \begin{bmatrix}
                (y_2 - y_1)(z_3 - z_1) - (z_2 - z_1)(y_3 - y_1) \\
                (z_2 - z_1)(x_3 - x_1) - (x_2 - x_1)(z_3 - z_1) \\
                (x_2 - x_1)(y_3 - y_1) - (y_2 - y_1)(x_3 - x_1)
            \end{bmatrix},\\
            \\
        D = -(K_1 x_1 + K_2 y_1 + K_3 z_1).
        \end{array} \\
    \end{aligned}
\end{equation}

In addition, the confidence value for each plane, denoted as \textit{probPlane}, is computed as the mean contact probability of the corresponding foot positions (extracted from the list \textit{P$_{contact}$}), using the $\operatorname{Mean()}$ function.

In the second phase, points in the terrain map\footnote{The map can be represented at different resolutions; see Section~\ref{results}.} are updated following the rules: 1) For points whose projections on the ground are not covered by the projections of any triangle\footnote{In the first phase, four triangles are constructed, each associated with one leg.}, the height information remains unchanged, 2) If a point's projection on the ground lies within at least one triangle, its height is updated via the $\operatorname{Update()}$ function, where the weighted position is computed considering the corresponding plane confidences.

In the final phase, the plane parameters are determined using the principal component analysis (PCA). Assuming there are $N$ points from the 2.5-D map projected beneath the robot's body, these $N$ points are combined into the point set $Points$. Then the parameters are calculated by
\begin{equation}
\begin{aligned}
\begin{array}{c}
    \mathbf{X} =
    \begin{bmatrix}
    x_1 - \bar{x} & y_1 - \bar{y} & z_1 - \bar{z} \\
    \vdots & \vdots & \vdots \\
    x_N - \bar{x} & y_N - \bar{y} & z_N - \bar{z}
    \end{bmatrix}, \\[8pt]
    \\
    \mathbf{C} = \frac{1}{N-1} \mathbf{X}^T \mathbf{X}, \\[8pt]
    \begin{bmatrix} K_1, K_2, K_3 \end{bmatrix} 
    = \bm{v}_{\min}, \\[8pt]
    D  = -(K_1 \bar{x} + K_2 \bar{y} + K_3 \bar{z}),
\end{array} 
\end{aligned}
\end{equation}
where $[x_i,y_i,z_i]$ is the position of $i$-th point in $Points$, $[\bar{x},\bar{y},\bar{z}]$ is the average position, $\mathbf{X}$ is the decentralized position matrix and $\mathbf{C}$ is the corresponding covariance matrix. $\bm{v}_{\min}$ is the eigenvector corresponding to the smallest eigenvalue of $\mathbf{C}$.

\begin{algorithm}
\caption{Terrain Estimation and Plane Update}
\label{TEalgorithm}
\KwIn{\textit{Terrain}, \textit{P$_{contact}$}, \textit{P$_{foot}$}}
\KwOut{\textit{Terrain}, \textit{Plane}}

\For{\textit{legIdx} $\gets 0$ \textbf{to} $3$}{
    \textit{Points} $\gets$ NeighborPoints(\textit{legIdx}, \textit{P$_{foot}$})\;
    \textit{locPlane}(\textit{legIdx}) $\gets$ ComputePlane(\textit{Points})\;
    \textit{probPlane}(\textit{legIdx}) $\gets$ Mean(\textit{P$_{contact}$}.at(\textit{Points}))\;
}

\For{\textit{Px} $\in$ \textit{Terrain.x}}{
    \For{\textit{Py} $\in$ \textit{Terrain.y}}{
        \textit{Pos} $\gets$ (\textit{Px}, \textit{Py})\;
        \textit{NumUnderPlanes} $\gets$ 0\;

        \For{\textit{Idx} $\gets 0$ \textbf{to} $3$}{
            \If{\textnormal{IsUnderPlane}(\textit{Pos},\textit{locPlane}(\textit{Idx}))}{
                \textit{NumUnderPlanes}++
            }
        }

        \If{\textit{NumUnderPlanes} = 0}{
            \textit{Terrain.at}(\textit{Pos}) $\gets$ \textit{Terrain.at}(\textit{Pos})\;
        }
        \Else{
            \textit{Terrain.at}(\textit{Pos}) $\gets$ Update(\textit{locPlane}, \textit{probPlane})\;
        }
    }
}
\textit{Plane} $\gets$ pointCloud2Plane(\textit{terrainEst})
\end{algorithm}

\subsection{Contact estimation}\label{sub_contact}
In quadrupedal locomotion, we define pseudo-contact as the situation in which physical foot-ground contact occurs, but the measured force falls below the detection threshold of force-based contact judgment. Such missed detections are detrimental to terrain and state estimation. To address this issue and to enhance the contact continuity over complex terrain, we propose a contact estimation method that fuses foot force measurements with estimated terrain information. For each leg, the contact probability is formulated as
\begin{equation}
P(c) = k_{\text{pos}} P(c)_{\text{pos}} + k_{\text{force}} P(c)_{\text{force}} , 
\label{eq:prob_combination}
\end{equation}
where $k_{[\cdot]}$ is the confidence weights of $[\cdot]$, and $P(c)_{[\cdot]}$ is the probability of contact from $[\cdot]$. $k_{\text{pos}}$ and $k_{\text{force}}$ are normalized such that $k_{\text{pos}}+k_{\text{force}}$=1.

$P(c)_{\text{pos}}$ is designed as an inverse sigmoid function, assigning higher probabilities to smaller distances, and vice versa. Specifically,
\begin{equation}
P(c)_{\text{pos}} = 1 - \delta\left(-\sigma_{\text{pos}} \left( 
\frac{K_1 x + K_2 y + K_3 z}{\sqrt{K_1^2 + K_2^2 + K_3^2}} 
- P_{\text{mid}} \right)\right), 
\label{eq:pos_prob}
\end{equation}
where $\sigma_{\text{pos}}$ is a scaling factor that controls the sensitivity of the sigmoid function to positional changes. $K_1$, $K_2$, and $K_3$ are the estimated plane parameters. $[x,y,z]^T$ denotes the 3D leg position. $P_{\text{mid}}$ is the midpoint of the sigmoid function.  $\delta$ is a sigmoid activation function that maps input values to the range $[0, 1]$ and is defined as
\begin{equation}
\delta(a) = \frac{1}{1 + e^{-a}}.
\label{eq:sigmoid}
\end{equation}

$P(c)_{\text{force}}$ is designed as a sigmoid function, assigning higher probabilities to larger forces, and vice versa:
\begin{equation}
P(c)_{\text{force}} = \delta\left(-\sigma_{\text{force}} (F - F_{\text{mid}})\right), 
\label{eq:force_prob}
\end{equation}
where $\sigma_{\text{force}}$ is the force scaling factor and $F_{\text{mid}}$ is the midpoint for the force sigmoid function, $F$ is the measured contact force\footnote{Note that we do not need to measure the directional contact force with a high-precision sensor. Instead, the proportional values from low-cost force sensors are enough.}.

\subsection{State estimation}\label{State_esti}

\subsubsection{State definition} For state estimation, we define the state variables as
\begin{equation}
\bm{X} = 
\begin{bmatrix}
\bm{p}_{\text{com}}, 
\bm{v}_{\text{com}},  
\bm{p}_{i} 
\end{bmatrix}^{\text{T}},
\vspace{-0.1cm}
\end{equation}
where $\bm{p}_{\text{com}}$ represents the robot body position, $\bm{v}_{\text{com}}$ represents the velocity of the body, and $\bm{p}_{i}$ represent the positions of the $i$-th foot. All states are represented in the world coordinate system.

The state dynamics are described by
\begin{align}\label{eq_state_1}
\begin{cases}
\dot{\bm{p}}_{\text{com}} = \bm{v}_{\text{com}}, \\
\dot{\bm{v}}_{\text{com}} = \bm{a}_{\text{com}} + \bm{g}, \\
\dot{\bm{p}}_{i} = 0 & \forall i \in \text{contact},\\
\dot{\bm{p}}_{i}\doteq\bm{v}_{\mathrm{com}}-\mathbf{J}_i \dot{\bm{q}}_i \quad & \forall i \notin  \text{contact}.
\end{cases}
\end{align}
where, $\mathbf{J}_{i}$ is the Jacobian matrix of the $i$-th leg, and $\bm{q}_{i}$ is the joint variable vector of the $i$-th leg.

Furthermore, we define the observation vector as
\begin{equation}
\bm{Z} = 
\begin{bmatrix}
\bm{p}_{\text{com}} - \bm{p}_{i} &
\bm{v} &
\bm{z}
\end{bmatrix}^{\text{T}},
\end{equation}
where $\bm{p}_{\text{com}} - \bm{p}_{i}$ represents the relative position of the body and the $i$-th foot, $\bm{v}$ represents the velocity vector of the feet in the body frame, and $\bm{z}$ represents the $z$-coordinate vector of the feet in the world frame.

The observation equation is formulated as
\begin{align}\label{eq_obs_1}
\begin{cases}
\bm{p}_{\text{com}} - \bm{p}_{i} = \mathbf{J}_{i} \cdot \bm{q}_{i}, \\
\bm{v}_{i} = {}^{o}\mathbf{R}_{B} \left( (\boldsymbol{\omega}_{OB}^{B})_{\times} \cdot {}^{B}_{i}\bm{p} + \frac{d {}^{B}_{i}\bm{p}}{dt} \right),
\\
z_i = \left(1-P(c)_{i}\right)\cdot \bm{p}_{i}[2]+P(c)_{i} \cdot z_{\text{est}\_i},\\ 
z_{\text{est}\_i} = \left( -K_1 \cdot \bm{p}_{i}[0] - K_2 \cdot \bm{p}_{i}[1] + D \right) / K_3,
\end{cases}
\end{align}
where the parameters are defined as follows:
\begin{itemize}
    \item ${}^{o}\mathbf{R}_{B}$ is the rotation matrix from the body frame to the world frame.
    \item $\boldsymbol{\omega}_{OB}^{B}$ is the angular velocity expressed in the robot's body frame.
    \item $(\cdot)_{\times}$ represents the skew-symmetric operator.
    \item $\bm{v}_{i}$ is the velocity vector of the $i$-th foot.
    \item ${}^{B}_{i}\bm{p}$ is the position vector of the $i$-th foot expressed in the body frame.
    \item $z_{i}$ is the vertical position of the $i$-th foot.
    \item $z_{\text{est}\_i}$ is the esitmated vertical position of the $i$-th foot.
    \item $P(c)_{i}$ is the contact probability of the $i$-th foot.
    \item $K_1$, $K_2$, $K_3$, $D$ are the plane parameters obtained from the terrain estimation module.
\end{itemize}

In contrast to \cite{yang2021a1qpmc}, this estimation method updates the leg height components in the measurement vector using the estimated plane parameters and contact probabilities.

\subsubsection{Kalman Filter (KF) formulation}
The predicted error covariance is computed as
\begin{equation}
    \mathbf{P_k}^{-} = \mathbf{A_k} \mathbf{P_{k-1}} \mathbf{A_k}^T + \mathbf{Q},
\end{equation}
where $\mathbf{Q}$ is the process noise covariance matrix. It is updated based on estimated contact, such that the foot-end velocity residuals of the swing leg are ignored. $\mathbf{A}_k$ is the discrete-time state transition matrix derived from~(\ref{eq_state_1}).

The estimated states are updated as
\begin{equation}
\begin{aligned}
    \mathbf{K_k} &= \mathbf{P_k}^{-} \mathbf{H}^T(\mathbf{H} \mathbf{P_k}^{-} \mathbf{H}^T + \mathbf{R})^{-1}, \\
    \mathbf{\bar{X}_k} &= \mathbf{\bar{X}_k}^{-} + \mathbf{K_k} \left( \mathbf{Z_k} - \mathbf{H} \mathbf{\bar{X}_k}^{-} \right), \\
    \mathbf{P_k} &= (\mathbf{I} - \mathbf{K_k} \mathbf{H}) \mathbf{P_k}^{-},
\end{aligned}
\end{equation}
where $\mathbf{R}$ is the measurement noise covariance matrix, $\mathbf{H}$ is the observation matrix derived from~(\ref{eq_obs_1}), $\mathbf{I}$ is the identity matrix and $\mathbf{P}_{k}$ denotes the state covariance. Similar to the process covariance matrix, $\mathbf{R}$ is adjusted to only offer high confidence when the leg is in contact.

\section{Safety guarantees in Mid-level control}\label{sec:safe}
To provide rigorous safety guarantees, we design constraints using CBFs. Specifically, by constraining the control input such that the state of the robot satisfies the inequality in \eqref{cbf_in}, the state can be ensured to remain within the safe set defined as $\mathcal{S} = \{ \bm{x} \in \mathbb{R}^n \mid h(\bm{x}) \geq 0 \}$.
\begin{equation}\label{cbf_in}
    \dot{h}(\bm{x}) + \alpha(h(\bm{x})) \geq 0,
\end{equation}
where $\alpha(\cdot)$ is a class $\kappa$ function.

To ensure both global safety (Subsection.~\ref{sec: global}) and local safety~(Subsection.~\ref{sec: local}), CBF-MPC is formulated as
\begin{equation}\label{MPC}
    \begin{aligned}
    \min_{\{\mathbf{X}[k], \mathbf{u}[k]\}} & \quad \sum_{k=0}^{N-1} \left( \|\mathbf{X}[k] - \mathbf{X}_{\text{ref}}[k]\|_{\mathbf{Q_w}}^2 + \|\mathbf{u}[k]\|_{\mathbf{R_w}}^2 \right), \\
    \text{s.t.} & \quad \mathbf{X}[k+1] = \mathbf{A}_d \mathbf{X}[k] + \mathbf{B}_d \mathbf{u}[k] + \mathbf{C}_d, \\
    & \quad \mathbf{X}[0] = \bm{x}_0, \\
    & \quad \mathbf{W}_i \mathbf{R}_i \mathbf{u}_i[k] \leq \mathbf{0}, \quad i = 1, \dots, 4, \\
    & \quad f_{\text{min}} \leq (\mathbf{R}_i \mathbf{u}_i[k])_z \leq f_{\text{max}}, \quad i = 1, \dots, 4, \\
    & \quad \mathbf{\delta}_i[k] \mathbf{u}_i[k] = \mathbf{0}, \quad i = 1, \dots, 4, \\
    & \quad \dot{h}_{\text{glob}}(\mathbf{X}[k]) + \alpha_{1}(h_{\text{glob}}(\mathbf{X}[k])) \geq 0 \\
    & \quad \dot{h}_{\text{local}}(\mathbf{X}[k]) + \alpha_{2}(h_{\text{local}}(\mathbf{X}[k])) \geq 0 \\
\end{aligned}
\end{equation}
where the parameters are defined as follows:
\begin{itemize}  
    \item $\mathbf{X} \in \mathbb{R}^{12\times N}$, $\mathbf{X}_{\text{ref}} \in \mathbb{R}^{12\times N}$, and $\mathbf{u} \in \mathbb{R}^{12\times N}$ are the state sequence, the reference sequence, and the ground reaction force within the prediction horizon, respectively.  
    \item $\mathbf{Q_w}$ is the state weighting matrix.  
    \item $\mathbf{R_w}$ is the control input weighting matrix.  
    \item $\mathbf{A}_d$ is the discrete-time state transition matrix.  
    \item $\mathbf{B}_d$ is the discrete-time control input matrix.  
    \item $\mathbf{C}_d$ is the gravity term.  
    \item $\bm{x}_0$ is the initial state.  
    \item $\mathbf{W}_i$ is the friction cone constraint matrix of leg $i$.  
    \item $\mathbf{R}_i$ is the rotation matrix mapping the ground reaction force of leg $i$ to the plane frame.  
    \item $f_{\text{min}}, f_{\text{max}}$ are the bounds on the ground reaction forces component along the plane normal direction.  
    \item $\mathbf{\delta}_i$ is the binary contact decision variable at leg $i$.  
    \item $h_{\text{glob}}(\cdot)$ is the global CBF-based constraint.  
    \item $h_{\text{local}}(\cdot)$ is the local CBF-based constraint.  
    \item $\alpha_{1}, \alpha_{2}$ are class $\kappa$ functions for corresponding constraints.  
\end{itemize} 

The details about the MPC formulation and solution can be found in \cite{ding2024robust}. Here, we focus on the CBF-based constraints.

\subsection{Global safety}\label{sec: global}
To achieve global safety, namely keep away from unsafe regions, we first identify potentially hazardous locations along the commanded motion direction using the 2.5-D map. Then, we define the CBF-based constraints. 

\begin{figure}
\centerline{\includegraphics[width=1\linewidth]{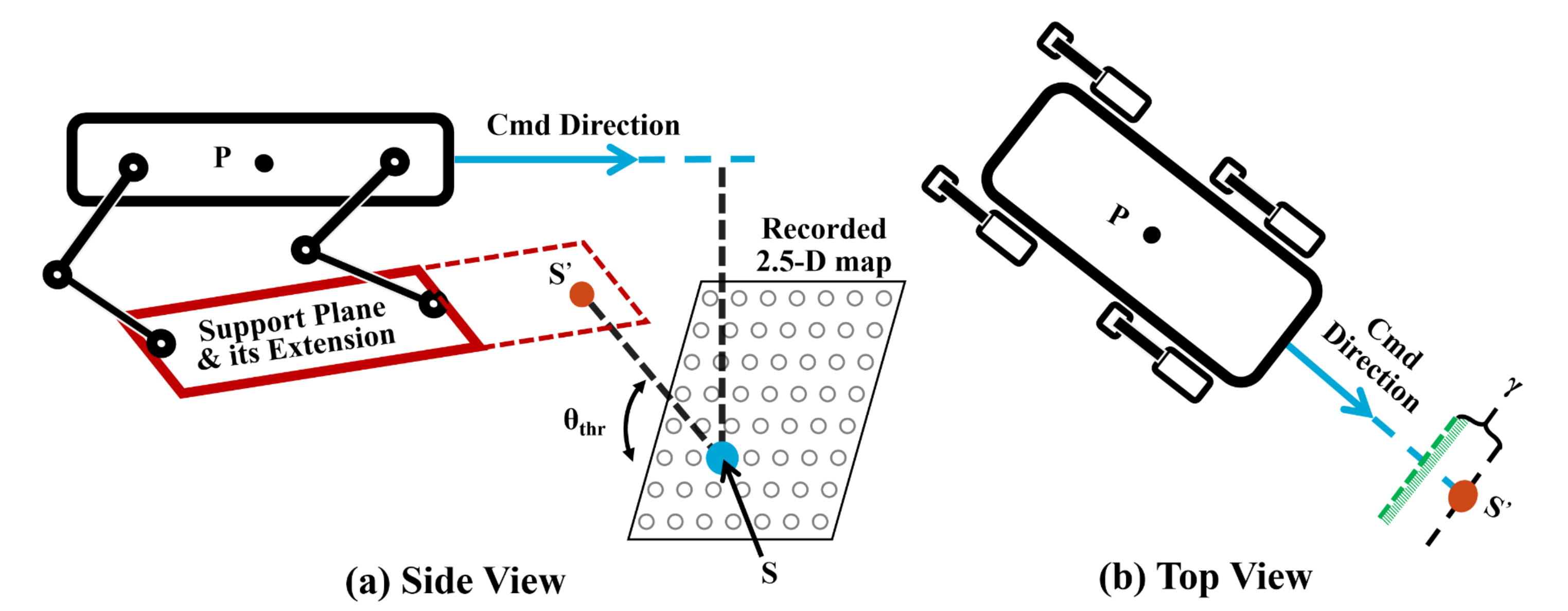}}
\vspace{-0.4cm}
\caption{Global safety CBF design. (a) Side view. (b) Top view. The gray point cloud represents previously recorded terrain points during locomotion. The green dotted line denotes the critical safety boundary. $\theta_{thr}$ is the angle relative to the horizontal plane and represents the maximum traversable slope of the robot.}
\label{global_cbf}
\vspace{-3mm}
\end{figure}

As illustrated in Fig.~\ref{global_cbf}(a), along the commanded (Cmd) motion direction, we search the recorded 2.5-D map to identify a potential hazardous point. Specifically, we determine the closest point $\bm{S}$ that satisfies the following conditions: the height difference between $\bm{S}$ and $\bm{P}$, $h_{\text{diff}}=\left| \bm{P}_\text{z}-\bm{S}_\text{z}\right|$, exceeds the predefined threshold $h_{\text{thr}}$, and the horizontal distance $l$ between $\bm{S}$ and the robot satisfies 
\begin{equation}
    l_{min}\leq l\leq l_{max}.
\end{equation} 
Here, $l_{min}$ and $l_{max}$ denote the minimum and maximum detection distances, which prevent overly close disturbances or distant terrain variations from being falsely identified as hazardous\footnote{The point $\bm{S}$ does not exist if no point on the 2.5-D map meets this requirement, indicating that no hazardous area is detected.}. 

Once the hazardous point $\bm{S}$ is identified, given the threshold slope angle $\theta_{\text{thr}}$, $\bm{S}$ is further projected onto the current support plane to obtain the projected point $\bm{S}'$.

Then, the safe set is defined as the region on the robot's side of the safety boundary (shown in Fig.~\ref{global_cbf}(b)) after $\gamma$ relaxation, mathematically expressed as:
\begin{equation}
    \mathcal{S} = \left\{ \bm{P_\text{xy}} \in \mathbb{R}^2 \mid \bm{d_\text{xy}} \cdot (\bm{S'_\text{xy}} - \bm{P_\text{xy}}) \geq \gamma \right\},
\end{equation}
where $\bm{P_\text{xy}}$, $\bm{d_\text{xy}}$, and $\bm{S'_\text{xy}}$ denote the projections of $\bm{P}$, $\bm{d}$, and $\bm{S'}$ onto the x-y plane, respectively. $\bm{d}$ is the direction vector marking the commanded walking direction, as shown in Fig.~\ref{global_cbf}(a) and (b).

Following the formulation in \eqref{cbf_in}, the CBF-based constraint of the above safe set is established as:
\begin{equation}
    -\bm{d_\text{xy}} \cdot \bm{\dot{P}_\text{xy}} \geq -\alpha_\text{glob} \left( \bm{d_\text{xy}} \cdot \bm{S'_\text{xy}} - \bm{d_\text{xy}} \cdot \bm{P_\text{xy}} - \gamma \right)
\end{equation}
where $\alpha_{\text{glob}}$ is a positive constant.

\subsection{Local safety}\label{sec: local}
\begin{figure}
\centerline{\includegraphics[width=1\linewidth]{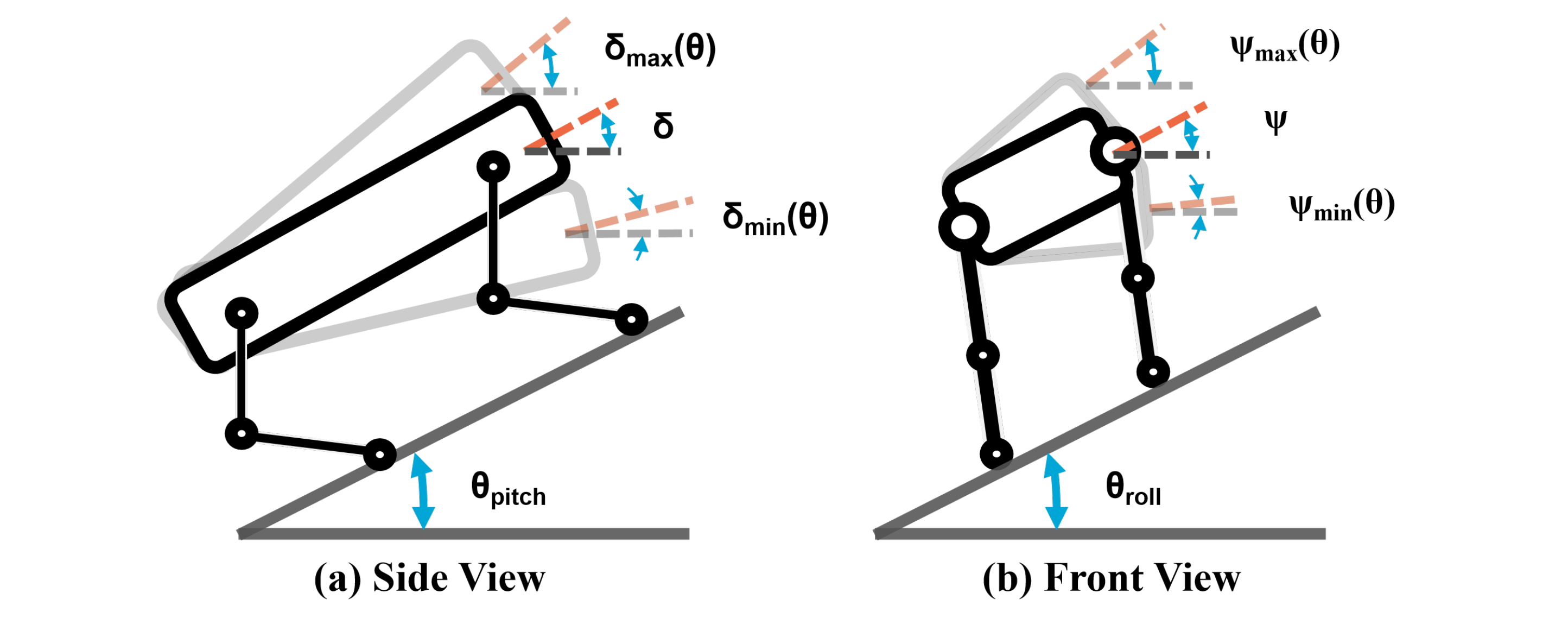}}
\vspace{-0.4cm}
\caption{Constrained body rotation for local safety in side and front views.}
\label{local_cbf}
\vspace{-5mm}
\end{figure}

\begin{figure*}
\vspace{-5mm} 
\centerline{\includegraphics[height=0.3\textheight,
    width=1.0\linewidth,
    keepaspectratio=false]{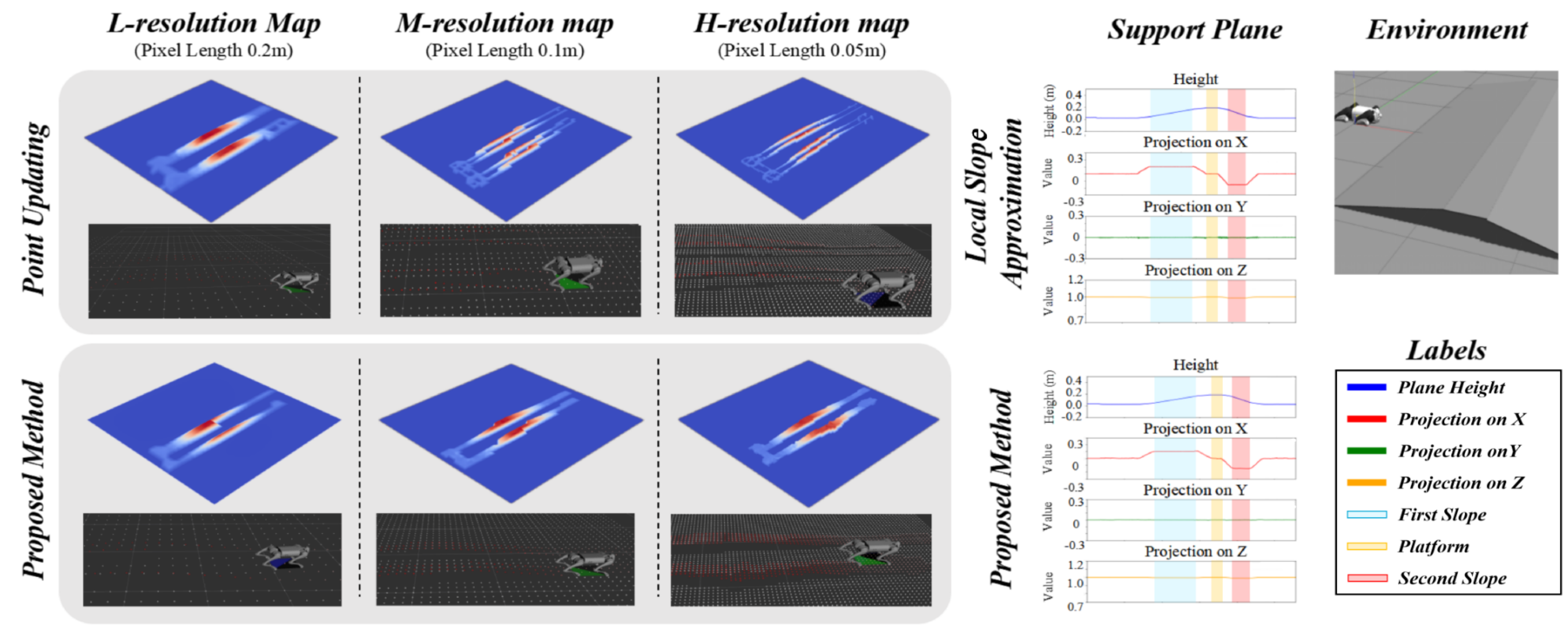}}
\vspace{-2mm}
\caption{The comparative experimental results of terrain estimation. For the 2.5-D map, we compare our method with the point updating method \cite{yang2023proprioception} when generating terrain maps with varying accuracy levels (low, medium, and high from left to right), corresponding to pixel lengths of 0.2m, 0.1m, and 0.05m, respectively. For the support plane, we compare our method with the local slope approximation strategy \cite{wang2023estimation}. The experiments are implemented in an environment with slopes and a platform.}
\label{terr_1}
\vspace{-3mm}
\end{figure*}
To guarantee local safety, we define the safe set for pitch and roll motions based on the estimated plane parameters, thereby adapting the body orientation to the terrain to mitigate body-terrain collisions. As illustrated in Fig.~\ref{local_cbf}. The safety set is expressed as:
\begin{equation}
    \begin{aligned}
        &\mathcal{S}_{\phi} = \left\{ \phi \in \mathbb{R} \mid \theta_{\text{pitch}} - \Delta \phi \leq \phi \leq \theta_{\text{pitch}} + \Delta \phi \right\}\\
        &\mathcal{S}_{\psi} = \left\{ \psi \in \mathbb{R} \mid \theta_{\text{roll}} - \Delta \psi \leq \psi \leq \theta_{\text{roll}} + \Delta \psi \right\}\\
    \end{aligned}
\end{equation}
where $\theta_{\text{pitch}}$ and $\theta_{\text{roll}}$ are the estimated slope angles, $\Delta \phi$ and $\Delta \psi$ are relaxation tolerances.

Four independent CBF-based constraints are established as the upper and lower bounds of the safe set, following
\begin{equation}
    \begin{aligned}
        \dot{\phi} &\geq -\alpha_{\text{local}} (\phi - \phi_{\min})\\
        -\dot{\phi} &\geq \alpha_{\text{local}} (\phi - \phi_{\max})\\
        \dot{\psi} &\geq -\alpha_{\text{local}} (\psi - \psi_{\min})\\
        -\dot{\psi} &\geq \alpha_{\text{local}} (\psi - \psi_{\max})\\
    \end{aligned},
\end{equation}
where $\alpha_{\text{local}}$ is a positive constant. The boundary values of the safe sets are defined as 
$\phi_{\min} = \theta_{\text{pitch}} - \Delta\phi$, 
$\phi_{\max} = \theta_{\text{pitch}} + \Delta\phi$, 
$\psi_{\min} = \theta_{\text{roll}} - \Delta\psi$, and 
$\psi_{\max} = \theta_{\text{roll}} + \Delta\psi$. The corresponding numerical parameter values are summarized in Table~\ref{tab:parameters2}.

\begin{table}
    \centering
    \caption{List of parameters used in safety guarantees}
    \vspace{-2mm}
    \begin{tabular}{ll|ll}
        \hline
        \textbf{Parameter} & \textbf{Value} & \textbf{Parameter} & \textbf{Value} \\ 
        \hline
        $h_{\text{thr}}$ & $0.5$ m & $\gamma$ & $0.15$ m \\ 
        $L_{\text{max}}$ & $1.2$ m & $\Delta \phi$ & $10^{\circ}$ \\ 
        $L_{\text{min}}$ & $0.5$ m & $\Delta \psi$ & $10^{\circ}$ \\ 
        $\theta_{\text{thr}}$ & $60^{\circ}$ & $\alpha_{\text{global}}$ & $1.5$ \\ 
        $\alpha_{\text{local}}$ & $0.5$ & & \\ 
        \hline
    \end{tabular}
    \label{tab:parameters2}
\end{table}
\vspace{-3mm}
\section{Results}\label{results}
This section presents experimental results. To validate the estimation module, four simulation experiments were conducted to evaluate terrain, contact, and state estimation separately. In addition, a real-world experiment was performed to assess terrain estimation. For safety guarantees, a simulation experiment was designed to evaluate the effectiveness of the CBF-based constraints. The parameters in Table~\ref{tab:parameters2} and Table~\ref{tab:parameters1} were determined through empirical tuning, with the selection process guided by estimation stability considerations and hardware constraints. All simulation experiments were conducted in Gazebo 11, while the real-world experiments were carried out in a physical environment consisting of slopes, flat planes, and steep cliffs. The Unitree Go1 robot was used as the experimental platform for all tests. The video of the experimental results is available at: \textcolor{blue}{\href{https://youtu.be/-0QsJLGNU6k}{[Link to Video]}}.

\begin{table}
    \centering
    \caption{List of parameters for contact estimation.}
     \vspace{-2mm}
    \begin{tabular}{ll|ll}
        \hline
        \textbf{Parameter} & \textbf{Value} & \textbf{Parameter} & \textbf{Value} \\ 
        \hline
        $k_{\text{pos}}$ & $0.40$ & $k_{\text{force}}$ & $0.60$ \\ 
        $\sigma_{\text{pos}}$ & $1.28$ & $P_{\text{mid}}$ & $0.01$ \\ 
        $\sigma_{\text{force}}$ & $0.78$ & $F_{\text{mid}}$ & $5.00$ \\ 
        \hline
    \end{tabular}
    \label{tab:parameters1}
    \vspace{-3mm}
\end{table}

\subsection{Terrain estimation performance}
As shown in Fig.~\ref{terr_1}, we evaluated the terrain estimation performance in an environment with varying slopes. The proposed method was compared with two baselines: the point updating method from \cite{yang2023proprioception} and the local slope approximation method from \cite{lin2020contact} and \cite{wang2023estimation}.

\begin{figure}
\vspace{-3mm}
\centerline{\includegraphics[width=\linewidth]{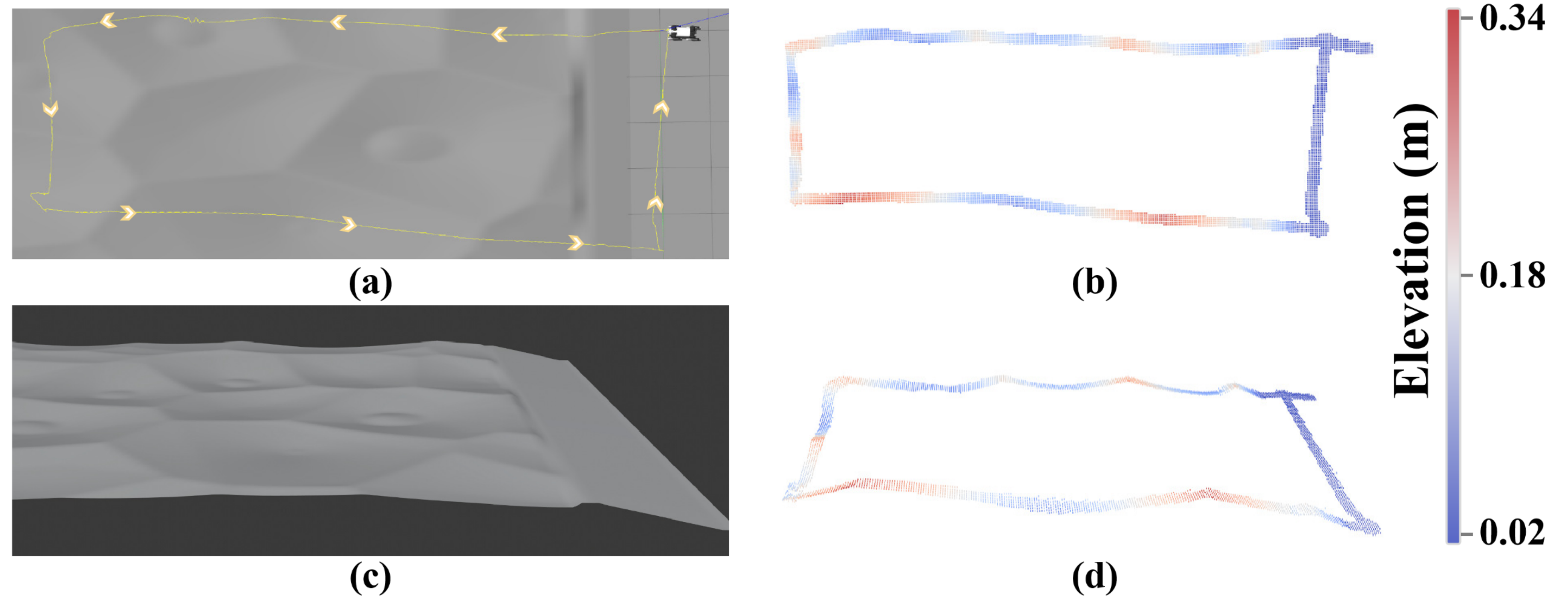}}
\vspace{-2mm}
\caption{Terrain estimation on unstructured terrain. (a) Motion trajectory; (b) top view of the 2.5-D map; (c) oblique side view of the terrain; (d) oblique side view of the 2.5-D map. (b) and (d) share the same colorbar.}
\label{uneven}
\vspace{-5mm}
\end{figure}

In terms of 2.5-D map generation, as the map resolution increases, the point updating method in \cite{yang2023proprioception} results in a noticeable trench-like gap between the point clouds updated by the left and right legs, whereas the proposed method consistently produces a smooth terrain reconstruction. Consequently, our approach eliminates the need for additional convolutional averaging filtering, ensuring a clearly delineated terrain representation. Regarding the estimation of the support plane, the proposed method achieves a consistent overall result compared with the local slope approximation \cite{wang2023estimation}. Quantitative experiment indicates that the proposed method achieves mean angular errors of $0.1399^\circ$ and $0.3320^\circ$ on the first and second slopes\footnote{The errors on the slopes are defined as the angle between the normal vectors of the predefined terrain model and the estimated plane.}, respectively, with a platform height error of -0.1987 cm. In contrast, the local slope approximation method yields mean errors of $0.1490^\circ$ and $0.2271^\circ$ on the two slopes, along with a platform height error of -0.1775 cm. Overall, the estimation errors of both methods remain at a low level, indicating accurate support plane estimation. Nevertheless, the proposed method further enables high-resolution 2.5-D map reconstruction, achieving unified terrain reconstruction and support plane estimation.

\begin{figure}
\vspace{-5mm}
\centerline{\includegraphics[height=0.27\textheight,
    width=1.0\linewidth,
    keepaspectratio=false]{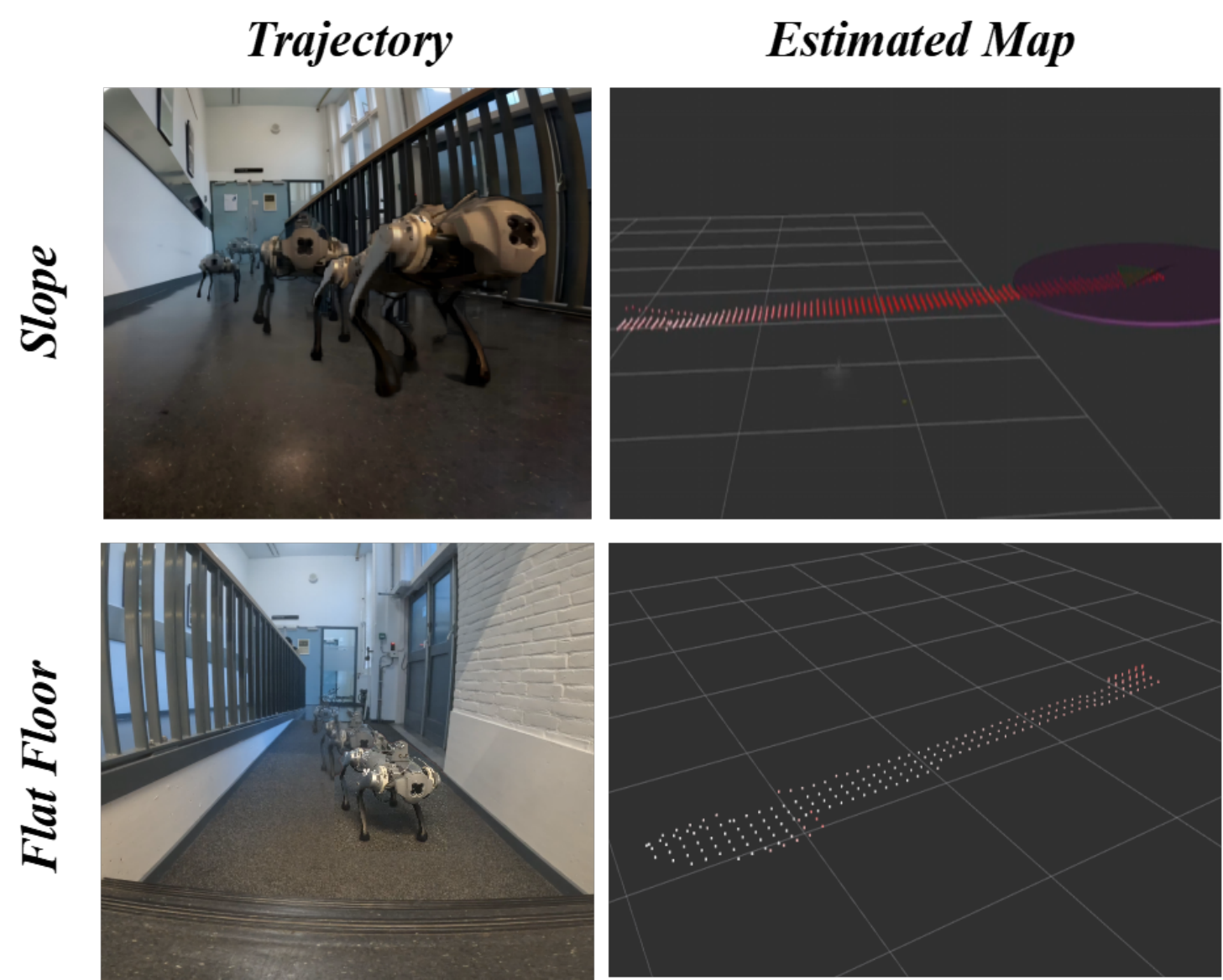}}
\vspace{-2mm}
\caption{Hardware experiments for terrain estimation when walking across a slope (top panel) and a flat ground (bottom panel).}
\label{terrain_real}
\vspace{-6mm}
\end{figure}

In addition, a qualitative experiment was conducted on unstructured terrain. As shown in Fig.~\ref{uneven}, the environment contains randomly generated irregular depressions and slopes. The robot traverses the environment along a closed-loop path. The estimated 2.5-D map effectively reflects the geometric characteristics and elevation variations of the terrain. The experimental results indicate that the proposed method maintains stable estimation performance on non-uniform and irregular surfaces.

To further validate the effectiveness of terrain estimation, we conducted a hardware experiment. As shown in Fig.~\ref{terrain_real}, the estimated point cloud on flat terrain remains close to zero elevation (white point in the bottom right image). On the slope, the estimated point cloud captures the terrain profile, i.e., the height of the point cloud increases as the robot moves forward. These results demonstrate that the proposed method achieves reliable terrain estimation when the robot operates on both flat and sloped surfaces, confirming its reliability in real-world deployments.

\vspace{-2mm}
\subsection{Contact estimation performance}

\begin{figure}
\centerline{\includegraphics[width=\linewidth]{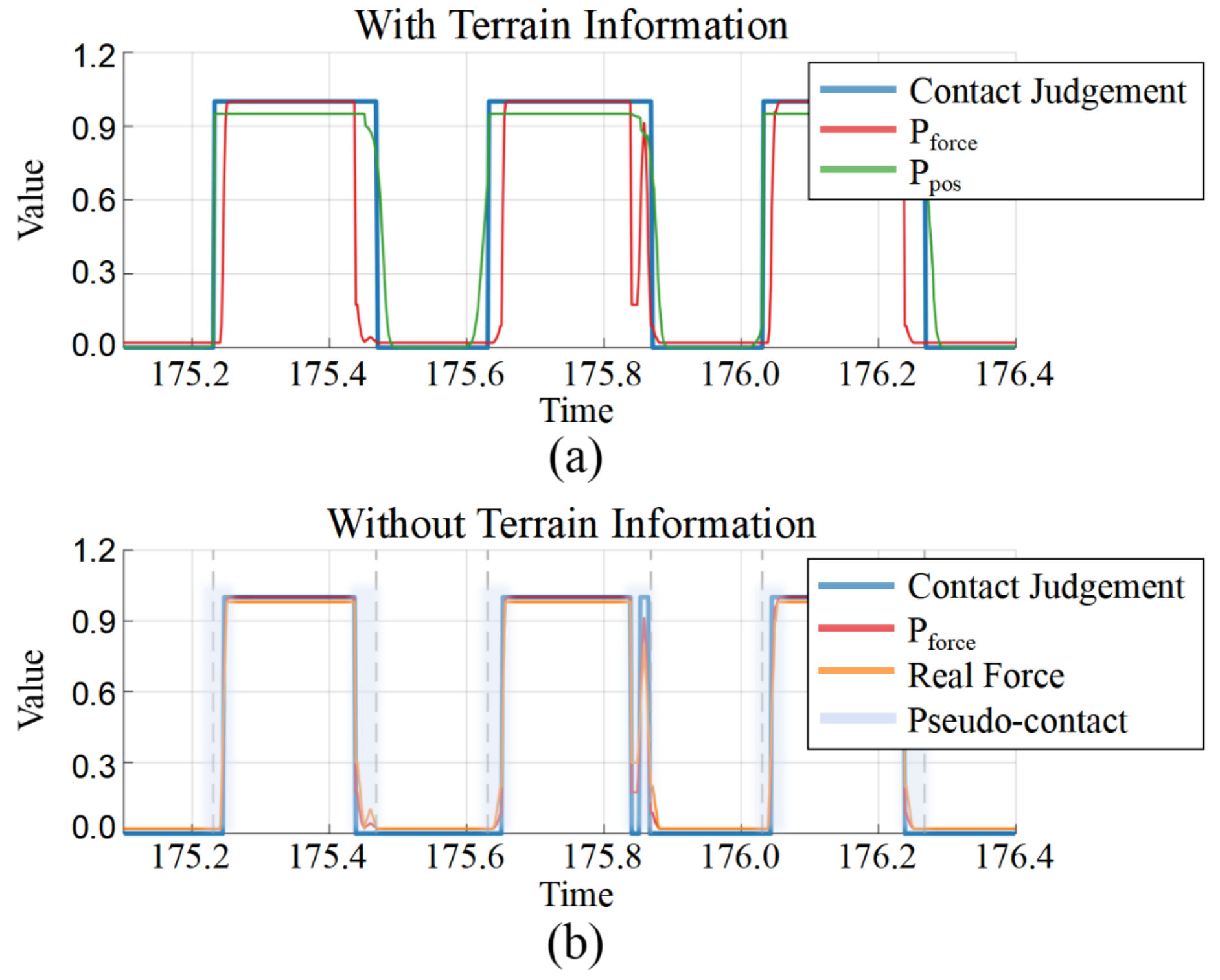}}
\vspace{-0.4cm}
\caption{Contact estimation with and without terrain information. Pseudo-contact refers to instances where contact occurs, but the measured force remains below the sensor threshold. For clarity, the contact judgment from (a) is overlaid in (b) as a dashed curve for direct comparison.}
\label{Contact}
\end{figure}

For the contact estimation, we use the parameters listed in Table~\ref{tab:parameters1}. As shown in Fig.~\ref{Contact}, the contact judgment with terrain information effectively handles pseudo-contact (light blue area in Fig.~\ref{Contact}(b)) compared to the contact judgment derived from only force measurements. Specifically, the contact judgment with terrain information covers pseudo-contact and maintains the continuity of contact in the presence of variations in contact force. 

\subsection{State estimation performance}

The performance of state estimation is evaluated by comparing the estimated state with the ground truth provided by the Gazebo simulator. In this scenario, the robot was commanded to walk in a scenario with multiple plane-slope switches, and ten repeatable experiments were conducted, each with and without terrain information. 

As mentioned in Section~\ref{State_esti}, the proposed method uses terrain information and contact probability to update leg heights. Since this change primarily influences the estimation of the CoM height, the experimental results presented here focus only on vertical motion.

\begin{figure}
\centerline{\includegraphics[height=0.25\textheight,
    width=1.0\linewidth,
    keepaspectratio=false]{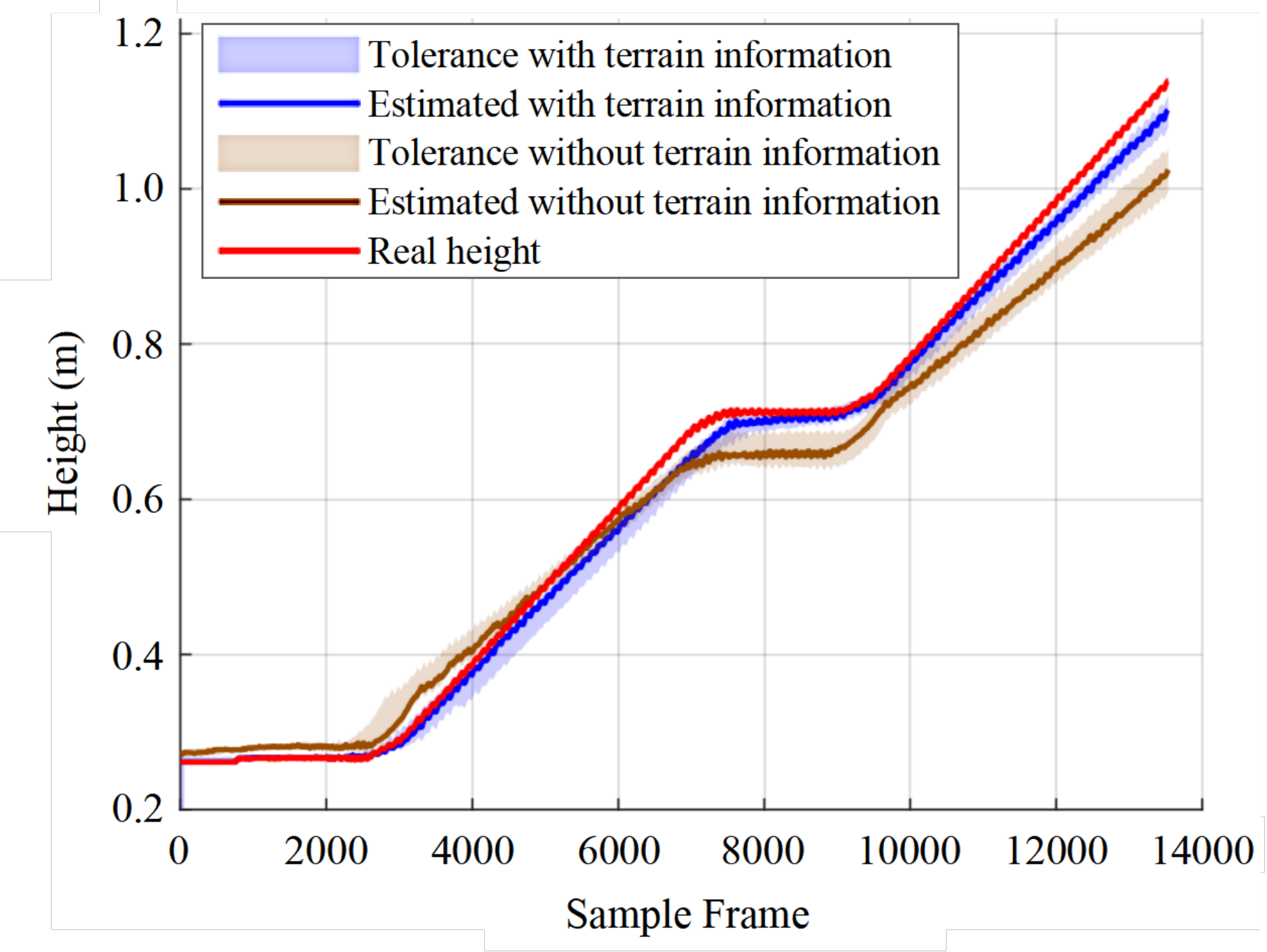}}
\vspace{-0.4cm}
\caption{Estimated CoM height profiles, with and without incorporating terrain information.}
\label{State}
\vspace{-5mm}
\end{figure}

The experimental results are reported in Fig.~\ref{State}. Analysis of the data showed that we reduced the mean absolute error (MAE) between state estimates and true values by 64.8\%, while also decreasing the error variance across experiments by 47.2\%. That is, incorporating terrain information effectively improves the accuracy of state estimation.

\subsection{Safety guarantee}

To evaluate the role of CBF constraints in safety guarantees, we designed a test environment with slopes and planes. The robot was first commanded to follow a predefined route to gather sufficient terrain information. Subsequently, it was directed toward the hazardous boundary with actions that could violate safety constraints.

\begin{figure}
\centerline{\includegraphics[width=1.0\linewidth]{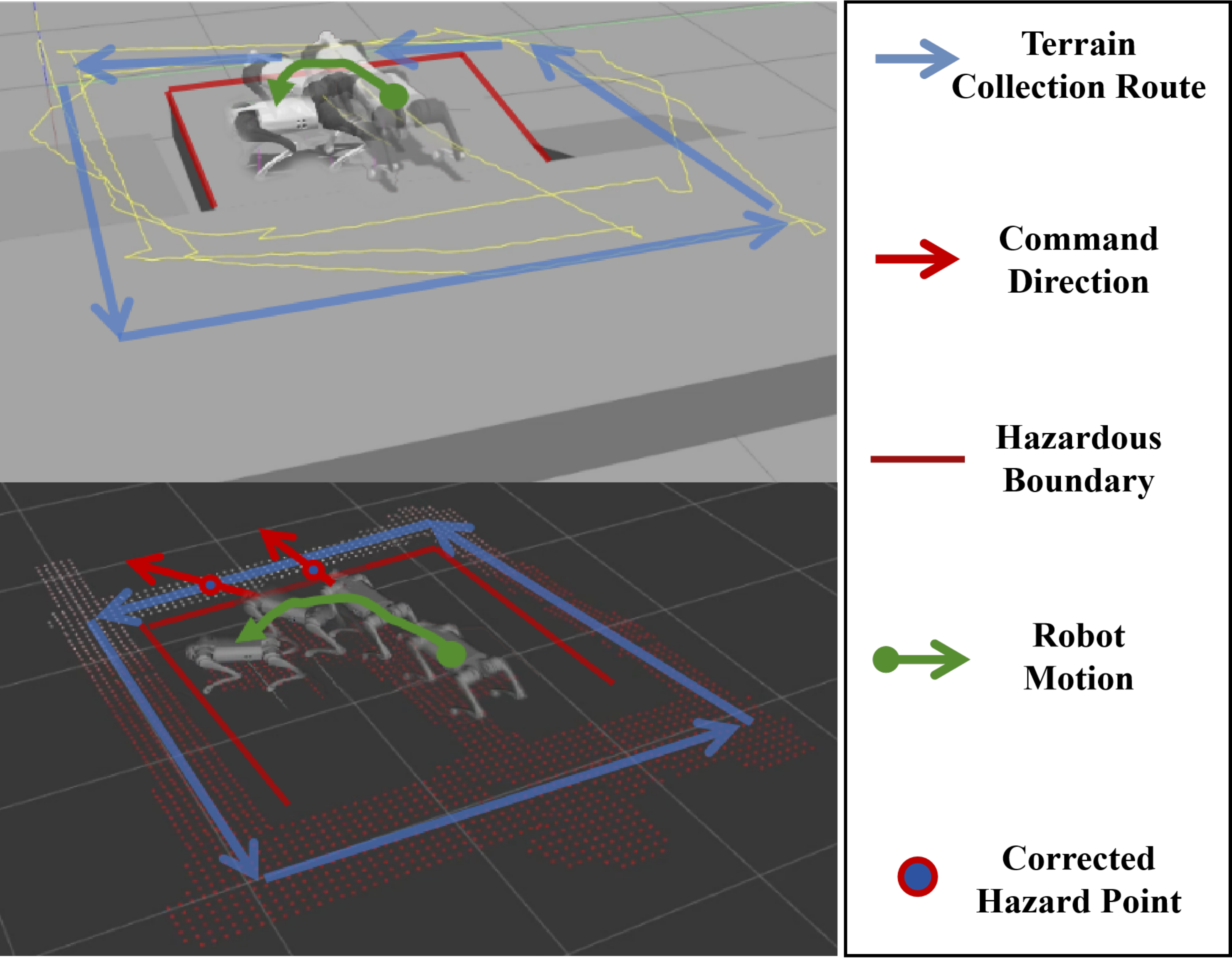}}
\vspace{-0.2cm}
\caption{An illustration of CBF constraints for avoidance in hazardous areas. The blue arrow represents the terrain collection route, the red arrow indicates the command direction, and the blue point marks the corrected hazard points. Upon approaching the hazardous boundary (highlighted by the red segment), the robot executes a stop-approaching maneuver. The continuous behavior is demonstrated in the supplementary video.}
\label{cbf_simu}
\vspace{-5mm}
\end{figure}

The results demonstrate that, without CBF constraints, the robot entered the hazardous region and finally tipped over (see the supplementary video). In contrast, with CBF constraints, the robot detected the hazardous points, activated a safety-critical boundary, and generated safe motions. Consequently, the robot successfully moved away from the hazardous region and prevented tipping over. 

The safe motion is visualized in Fig.~\ref{cbf_simu}. Note that after stopping on the front edge, we commanded the robot to turn left. With the CBF-MPC approach, the robot could move again until it reached the left hazardous boundary. The resultant CoM trajectory is plotted by the green curve.

\section{Limitations and future work}\label{Limitations}

Although the proposed framework enables safe locomotion under degraded external perception conditions, such as poor lighting or smoke, the perception range of the current work is limited to the local region and areas covered by the historical trajectory. Consequently, the method can provide safety guarantees only within explored regions and cannot impose anticipatory constraints on unknown areas.

Notably, the 2.5-D map constructed by the estimation module is compatible with external perception modalities. In future work, we plan to integrate our method with external perception-based simultaneous localization and mapping (SLAM) methods to provide critical safety redundancy for the system in cases of illumination degradation, smoke, or dust occlusion.
\vspace{-3mm}
\section{Conclusion}\label{conclusion}
This work achieves safe locomotion for quadrupedal robots relying solely on proprioceptive sensing. Specifically, we start with a novel terrain estimation method and then couple the terrain information with state and contact estimation. Incorporating the estimation into the MPC framework, we further introduce CBF constraints to provide safety guarantees at both global and local levels. Experiments have demonstrated that the coupled estimation system contributes to more accurate estimation, and the CBF constraints effectively guarantee safety.

\section*{Acknowledgement}
We thank Fabio Boekel, Weilin Xia, and Qifan Luo from TU Delft for their valuable contributions to this work.

\bibliographystyle{IEEEtran}
\bibliography{IEEEabrv, references}

@article{ding2024robust,
  title={Robust quadrupedal jumping with impact-aware landing: Exploiting parallel elasticity},
  author={Ding, Jiatao and Atanassov, Vassil and Panichi, Edoardo and Kober, Jens and Della Santina, Cosimo},
  journal={IEEE Transactions on Robotics},
volume={40},
pages={3212--3231},
  year={2024}
}

@article{gangapurwala2022rloc,
  title={Rloc: Terrain-aware legged locomotion using reinforcement learning and optimal control},
  author={Gangapurwala, Siddhant and Geisert, Mathieu and Orsolino, Romeo and Fallon, Maurice and Havoutis, Ioannis},
  journal={IEEE Transactions on Robotics},
  volume={38},
  number={5},
  pages={2908--2927},
  year={2022},
  publisher={IEEE}
}

@article{shi2023terrain,
  title={Terrain-aware quadrupedal locomotion via reinforcement learning},
  author={Shi, Haojie and Zhu, Qingxu and Han, Lei and Chi, Wanchao and Li, Tingguang and Meng, Max Q-H},
  journal={arXiv preprint arXiv:2310.04675},
  year={2023}
}

@inproceedings{wermelinger2016navigation,
  title={Navigation planning for legged robots in challenging terrain},
  author={Wermelinger, Martin and Fankhauser, P{\'e}ter and Diethelm, Remo and Kr{\"u}si, Philipp and Siegwart, Roland and Hutter, Marco},
  booktitle={2016 IEEE/RSJ International Conference on Intelligent Robots and Systems},
  pages={1184--1189},
  year={2016},
  organization={IEEE}
}

@inproceedings{norby2022quad,
  title={Quad-SDK: Full stack software framework for agile quadrupedal locomotion},
  author={Norby, Joseph and Yang, Yanhao and Tajbakhsh, Ardalan and Ren, Jiming and Yim, Justin K and Stutt, Alexandra and Yu, Qishun and Flowers, Nikolai and Johnson, Aaron M},
  booktitle={ICRA Workshop on Legged Robots},
  year={2022}
}

@article{kweon1992high,
  title={High-resolution terrain map from multiple sensor data},
  author={Kweon, In-So and Kanade, Takeo},
  journal={IEEE Transactions on Pattern Analysis and Machine Intelligence},
  volume={14},
  number={2},
  pages={278--292},
  year={1992}
}

@article{jiang2019simultaneous,
  title={A simultaneous localization and mapping (SLAM) framework for 2.5 D map building based on low-cost LiDAR and vision fusion},
  author={Jiang, Guolai and Yin, Lei and Jin, Shaokun and Tian, Chaoran and Ma, Xinbo and Ou, Yongsheng},
  journal={Applied sciences},
  volume={9},
  number={10},
  pages={2105},
  year={2019},
  publisher={MDPI}
}

@inproceedings{belter2012estimating,
  title={Estimating terrain elevation maps from sparse and uncertain multi-sensor data},
  author={Belter, Dominik and {\L}abcki, Przemys{\l}aw and Skrzypczy{\'n}ski, Piotr},
  booktitle={2012 IEEE International Conference on Robotics and Biomimetics},
  pages={715--722},
  year={2012},
  organization={IEEE}
}

@article{fankhauser2016universal,
  title={A universal grid map library: Implementation and use case for rough terrain navigation},
  author={Fankhauser, P{\'e}ter and Hutter, Marco},
  journal={Robot Operating System The Complete Reference (Volume 1)},
  pages={99--120},
  year={2016},
  publisher={Springer}
}

@article{xie2024real,
  title={Real-time Support Terrain Mapping and Terrain Adaptive Local Planning for Quadruped Robots},
  author={Xie, Han and Cui, Chang and Zhong, Xunyu and Zhong, Xungao and Liu, Qiang},
  journal={IEEE Robotics and Automation Letters},
  volume={9},
  number={12},
pages={11018--11025},
year={2024},
  publisher={IEEE}
}

@article{fankhauser2018probabilistic,
  title={Probabilistic terrain mapping for mobile robots with uncertain localization},
  author={Fankhauser, P{\'e}ter and Bloesch, Michael and Hutter, Marco},
  journal={IEEE Robotics and Automation Letters},
  volume={3},
  number={4},
  pages={3019--3026},
  year={2018},
  publisher={IEEE}
}

@inproceedings{bledt2018cheetah,
  title={Mit cheetah 3: Design and control of a robust, dynamic quadruped robot},
  author={Bledt, Gerardo and Powell, Matthew J and Katz, Benjamin and Di Carlo, Jared and Wensing, Patrick M and Kim, Sangbae},
  booktitle={2018 IEEE/RSJ International Conference on Intelligent Robots and Systems},
  pages={2245--2252},
  year={2018},
  organization={IEEE}
}

@inproceedings{lin2020contact,
  title={Contact surface estimation via haptic perception},
  author={Lin, Hsiu-Chin and Mistry, Michael},
  booktitle={2020 IEEE International Conference on Robotics and Automation},
  pages={5087--5093},
  year={2020},
  organization={IEEE}
}

@inproceedings{wang2023estimation,
  title={Estimation of Ground Posture Angle for Quadruped Robots Based on IMU},
  author={Wang, Jiliang and Pan, Zheng and Niu, Zhihua and Liu, Shaoxun and Zhou, Shiyu and Wang, Rongrong},
  booktitle={2023 6th International Conference on Electronics Technology},
  pages={1269--1275},
  year={2023},
  organization={IEEE}
}

@inproceedings{yang2023proprioception,
  title={Proprioception and tail control enable extreme terrain traversal by quadruped robots},
  author={Yang, Yanhao and Norby, Joseph and Yim, Justin K and Johnson, Aaron M},
  booktitle={2023 IEEE/RSJ International Conference on Intelligent Robots and Systems},
  pages={735--742},
  year={2023},
  organization={IEEE}
}

@inproceedings{lee2024safety,
  title={Safety-critical control of quadrupedal robots with rolling arms for autonomous inspection of complex environments},
  author={Lee, Jaemin and Kim, Jeeseop and Ubellacker, Wyatt and Molnar, Tamas G and Ames, Aaron D},
  booktitle={2024 IEEE International Conference on Robotics and Automation},
  pages={3485--3491},
  year={2024},
  organization={IEEE}
}

@inproceedings{grandia2021multi,
  title={Multi-layered safety for legged robots via control barrier functions and model predictive control},
  author={Grandia, Ruben and Taylor, Andrew J and Ames, Aaron D and Hutter, Marco},
  booktitle={2021 IEEE International Conference on Robotics and Automation},
  pages={8352--8358},
  year={2021},
  organization={IEEE}
}

@article{shamsah2023integrated,
  title={Integrated task and motion planning for safe legged navigation in partially observable environments},
  author={Shamsah, Abdulaziz and Gu, Zhaoyuan and Warnke, Jonas and Hutchinson, Seth and Zhao, Ye},
  journal={IEEE Transactions on Robotics},
  volume={39},
  number={6},
  pages={4913--4934},
  year={2023},
  publisher={IEEE}
}

@inproceedings{unlu2024control,
  title={A control barrier function-based motion planning scheme for a quadruped robot},
  author={Unlu, Halil Utku and Gon{\c{c}}alves, Vinicius Mariano and Chaikalis, Dimitris and Tzes, Anthony and Khorrami, Farshad},
  booktitle={2024 IEEE International Conference on Robotics and Automation},
  pages={12172--12178},
  year={2024},
  organization={IEEE}
}

@inproceedings{saradagi2024body,
  title={Body-aware Local Navigation for Asymmetric Holonomic Robots using Control Barrier Functions},
  author={Saradagi, Akshit and Fredriksson, Scott and Koval, Anton and Nikolakopoulos, George},
  booktitle={2024 European Control Conference},
  pages={968--973},
  year={2024},
  organization={IEEE}
}

@inproceedings{schneider2024learning,
  title={Learning risk-aware quadrupedal locomotion using distributional reinforcement learning},
  author={Schneider, Lukas and Frey, Jonas and Miki, Takahiro and Hutter, Marco},
  booktitle={2024 IEEE International Conference on Robotics and Automation},
  pages={11451--11458},
  year={2024},
  organization={IEEE}
}

@article{brunke2022safe,
  title={Safe learning in robotics: From learning-based control to safe reinforcement learning},
  author={Brunke, Lukas and Greeff, Melissa and Hall, Adam W and Yuan, Zhaocong and Zhou, Siqi and Panerati, Jacopo and Schoellig, Angela P},
  journal={Annual Review of Control, Robotics, and Autonomous Systems},
  volume={5},
  number={1},
  pages={411--444},
  year={2022},
  publisher={Annual Reviews}
}

@inproceedings{bledt2018contact,
  title={Contact model fusion for event-based locomotion in unstructured terrains},
  author={Bledt, Gerardo and Wensing, Patrick M and Ingersoll, Sam and Kim, Sangbae},
  booktitle={2018 IEEE International Conference on Robotics and Automation},
  pages={4399--4406},
  year={2018},
  organization={IEEE}
}

@misc{yang2021a1qpmc,
  author       = {Shuo Yang},
  title        = {A1-QP-MPC-Controller: Quadruped MPC Controller},
  year         = {2021},
  howpublished = {\url{https://github.com/ShuoYangRobotics/A1-QP-MPC-Controller}},
  note         = {Accessed: 2024-11-30}
}

@inproceedings{li2023autonomous,
  title={Autonomous and Safety-Critical Stair Climbing via Nonlinear Model Predictive Control for Quadrupedal Robots},
  author={Li, Chengzhuo and Peng, Xiafu and Lan, Weiyao and Yu, Xiao},
  booktitle={2023 IEEE International Conference on Robotics and Biomimetics},
  pages={1--6},
  year={2023},
  organization={IEEE}
}

@article{panichi2025fly,
  title={On-the-Fly Jumping With Soft Landing: Leveraging Trajectory Optimization and Behavior Cloning},
  author={Panichi, Edoardo and Ding, Jiatao and Atanassov, Vassil and Yang, Peiyu and Kober, Jens and Pan, Wei and Della Santina, Cosimo},
  journal={IEEE/ASME Transactions on Mechatronics},
  year={2025},
  volume={30},
  number={4},
  pages={3142-3151},
  publisher={IEEE}
}

@article{hoeller2024anymal,
  title={Anymal parkour: Learning agile navigation for quadrupedal robots},
  author={Hoeller, David and Rudin, Nikita and Sako, Dhionis and Hutter, Marco},
  journal={Science Robotics},
  volume={9},
  number={88},
  pages={eadi7566},
  year={2024},
  publisher={American Association for the Advancement of Science}
}

@ARTICLE{11045071,
  author={Wang, Yu and Liu, Yufeng and Chen, Lingxu and Chen, Haoyao and Zhang, Shiwu},
  journal={IEEE Robotics and Automation Letters}, 
  title={Degradation-Aware LiDAR-Thermal-Inertial SLAM}, 
  year={2025},
  volume={10},
  number={8},
  pages={8035-8042},
  doi={10.1109/LRA.2025.3581127}}

@article{dhrafani2025firestereo,
  title={Firestereo: Forest infrared stereo dataset for uas depth perception in visually degraded environments},
  author={Dhrafani, Devansh and Liu, Yifei and Jong, Andrew and Shin, Ukcheol and He, Yao and Harp, Tyler and Hu, Yaoyu and Oh, Jean and Scherer, Sebastian},
  journal={IEEE Robotics and Automation Letters},
  year={2025},
  publisher={IEEE}
}

@article{he2024agile,
  title={Agile but safe: Learning collision-free high-speed legged locomotion},
  author={He, Tairan and Zhang, Chong and Xiao, Wenli and He, Guanqi and Liu, Changliu and Shi, Guanya},
  journal={arXiv preprint arXiv:2401.17583},
  year={2024}
}

@article{ding2025versatile,
  title={Versatile, robust, and explosive locomotion with rigid and articulated compliant quadrupeds},
  author={Ding, Jiatao and Yang, Peiyu and Boekel, Fabio and Kober, Jens and Pan, Wei and Saveriano, Matteo and Della Santina, Cosimo},
  journal={arXiv preprint arXiv:2504.12854},
  year={2025}
}

@article{atanassov2024curriculum,
  title={Curriculum-based reinforcement learning for quadrupedal jumping: A reference-free design},
  author={Atanassov, Vassil and Ding, Jiatao and Kober, Jens and Havoutis, Ioannis and Della Santina, Cosimo},
  journal={IEEE Robotics \& Automation Magazine},
  volume={32},
  number={2},
  pages={35--48},
  year={2024},
  publisher={IEEE}
}

\end{document}